\documentclass[sigconf,nonacm]{acmart}

\AtBeginDocument{%
  \providecommand\BibTeX{{%
    \normalfont B\kern-0.5em{\scshape i\kern-0.25em b}\kern-0.8em\TeX}}}

\begin{document}

\title{
An Aerial Manipulator for Perception-Driven Flower Targeting Toward Contactless Pollination in Vertical Farming
}

\author{Chenzhe Jin}
\affiliation{
\department{Department of Computer Science}
\institution{University College London}
\city{London}
\country{United Kingdom}}

\author{Zhuohang Wu}
\affiliation{
\department{Department of Computer Science}
\institution{University College London}
\city{London}
\country{United Kingdom}}

\author{Yifan Cai}
\affiliation{
\department{Department of Computer Science}
\institution{University College London}
\city{London}
\country{United Kingdom}}

\author{Xiangqi Li}
\affiliation{
\department{Department of Computer Science}
\institution{University College London}
\city{London}
\country{United Kingdom}}

\author{Jan Ming Kevin Tan}
\affiliation{
\department{Department of Computer Science}
\institution{University College London}
\city{London}
\country{United Kingdom}}

\author{Narsimlu Kemsaram}
\affiliation{
\department{Department of Artificial Intelligence}
\institution{University of Malaya}
\city{Kuala Lumpur}
\country{Malaysia}}

\author{Valerio Modugno}
\affiliation{
\department{Department of Computer Science}
\institution{University College London}
\city{London}
\country{United Kingdom}}


\begin{abstract}
  The decline of natural pollinators has created a major challenge for crop production in controlled indoor agriculture, particularly in vertical farming environments where natural insect pollination is absent. This motivates the development of robotic systems capable of performing precise flower-targeting tasks while minimizing physical interference with delicate floral structures.
  This paper presents an aerial manipulator platform for perception-driven flower detection, localization, and approach in vertical farming environments. The proposed system integrates onboard RGB-D-based perception, model predictive path integral (MPPI)-based unmanned aerial vehicle (UAV) control on a PX4 platform, and a lightweight 2-DoF manipulator for precise end-effector positioning. The platform is evaluated in both MuJoCo simulation and UAV lab experiments using a flower-targeting testbed.
  The experimental results demonstrate stable UAV flight, reliable flower localization, and centimeter-level end-effector positioning accuracy. In simulation, the proposed controller achieves consistent trajectory convergence and accurate target alignment. In the real-world UAV lab environment, the integrated perception–control–manipulation framework enables stable flower-targeted positioning and end-effector alignment under constrained aerial operation.
  These results validate the proposed aerial manipulator as a robust robotic carrier and positioning framework for future contactless pollination systems. While the current study focuses on perception-guided targeting and positioning, the developed platform provides a practical foundation for integrating advanced contactless end-effectors, including acoustic-based pollen manipulation modules, in future work.
\end{abstract}



\begin{teaserfigure}
  \includegraphics[width=\textwidth]{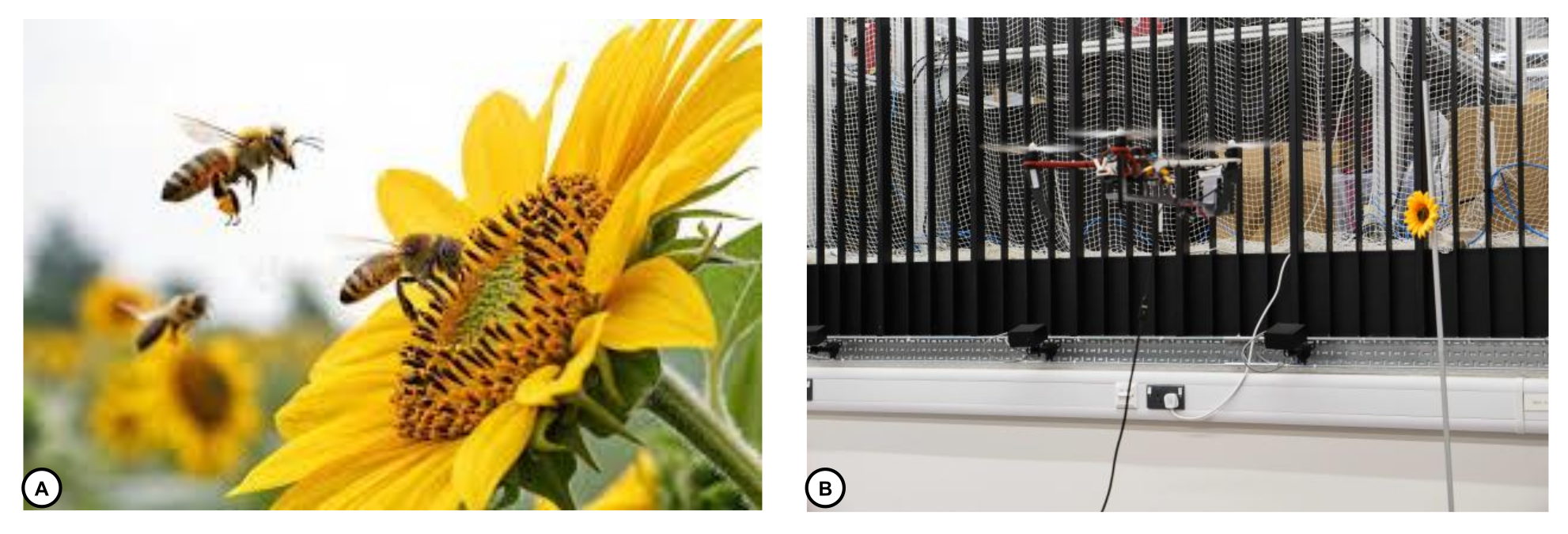}
  \caption{
  Conceptual illustration of pollination mechanisms: (A) Natural pollination by bees, and (B) An aerial manipulation platform toward future contactless pollination in vertical farming environments.
  }
  \Description{Enjoying the baseball game from the third-base
  seats. Ichiro Suzuki preparing to bat.}
  \label{fig:BlockDiagram}
\end{teaserfigure}

\maketitle

\section{Introduction}


The global decline of natural pollinators poses a significant threat to agricultural productivity and food security \cite{simpson2025review}. With the global population projected to reach nearly 10 billion by 2050, modern agriculture must adopt innovative technologies to ensure stable crop yields \cite{UNDESA_WPP_2022_Summary}, \cite{Mirzabaev2023_ClimateRiskMgmt}. Vertical farming has emerged as a promising solution, offering controlled environments for year-round production. However, the lack of natural insect pollination in these indoor facilities necessitates the development of autonomous robotic alternatives \cite{UNWaterFAO2024WaterStress642}.
Existing robotic pollination methods primarily rely on physical contact, such as mechanical brushes, vibrating sticks, or micro-drones equipped with ionic liquid gels \cite{chechetka2017materially}, \cite{broussard2023artificial}. While functional, these "contact-based" approaches carry a high risk of damaging delicate floral structures, such as petals and stigmas, which can paradoxically reduce the overall crop quality. To address these challenges, there is a growing demand for high-precision robotic systems capable of navigating the confined, dense canopies of vertical farms and delivering pollination agents with minimal physical interference (see Figure \ref{fig:BlockDiagram}).

In this paper, we present an aerial manipulator platform for perception-driven flower detection, localization, and approach in vertical farming environments. The system integrates onboard RGB-D perception, MPPI-based UAV control, and a lightweight 2-DoF manipulator to achieve stable flight and centimeter-level end-effector positioning near flower targets. It also establishes a practical platform for future integration of contactless pollination end-effectors.
The core of our platform lies in its integrated perception and control framework. We combine a real-time deep learning model for precise flower localization with a feedback MPPI controller \cite{belvedere2025feedback}. This synergy ensures stable flight dynamics and precise trajectory tracking even in the presence of the aerodynamic disturbances common in indoor farming. 
While the platform is designed to support various pollination tools, we demonstrate its potential through a prototype interface that paves the way for contactless acoustic manipulation—a promising technique to be explored in our future research to eliminate physical contact entirely via acoustic radiation pressure \cite{kemsaram2025acoustobots}, \cite{becsevli2025sonarios}, \cite{kemsaram2025cooperative}.





The main contributions of this work are three-fold: 1) System design: We designed a lightweight, high-precision aerial-manipulator platform optimized for the constrained spatial requirements of vertical farms, 2) Integrated control: We developed a robust control pipeline that leverages deep-learning-based perception and an advanced MPPI controller to achieve stable and accurate end-effector positioning near floral targets, and 3) Experimental validation: We conducted comprehensive evaluations in both MuJoCo simulations and UAV lab environments, demonstrating that the platform provides a reliable and scalable framework for future contactless pollination methods.

The remainder of this paper is organized as follows. Section 2 reviews related work. Section 3 presents the proposed aerial manipulator platform and overall architecture. Section 4 describes the system design, modeling, and control framework. Section 5 introduces the ROS 2-based system implementation. Section 6 presents simulation and real-world experimental evaluation. Finally, Section 7 concludes the paper and outlines future research directions.
\section{Literature Review}

This section reviews literature on vertical farming, its challenges, and robotic and drone-based pollination methods. These studies establish the context and highlight the research gap this paper aims to address.

\subsection{Vertical Farming}

The proposal of the vertical agriculture mode aims to increase agricultural land area through "upward construction", thereby maximizing yield \cite{ahlfeldt2023skyscraper}. 
It performs well in terms of water consumption and is attractive in countries with severe pollution and poor soil conditions \cite{verticalfarmingtemp1}.
These systems often utilize automatic irrigation and nutrient supply mechanisms \cite{verticalfarmingtemp2}. 
Its typical features include: innovative use of recycled water, supplemented by rainwater or water from seawater desalination plants, automatic air temperature and humidity control, solar panel lighting and heating, and adjustable 24-hour LED lighting \cite{verticalfarmingtemp1}. In addition, vertical agriculture is beneficial for growing various crops, including exotic and special plants that may not grow in traditional outdoor environments \cite{mir2022vertical}. These multiple responses not only meet the market demand for characteristics, but also promote agricultural biodiversity. The cultivation of various crops in a vertical stacking system is crucial for long-term food security and the ability to resist the impacts of disease and climate change \cite{verticalfarmingtemp5}.
However, vertical agriculture currently faces many challenges, such as pollination. It is the most tedious process in agriculture. More than 85\% of flowering plants rely on biological pollination, mainly through insect transmission, while the rest rely on non biological pollen transmission, mainly through wind pollination \cite{verticalfarmingtemp4}. For example, as mentioned in \cite{verticalfarmingtemp2}, crops that require insect pollination are at a disadvantage and require workers to spend time and effort on manual pollination. The introduction of modern agricultural technology, including sensor technology, robot technology, artificial intelligence, etc., has promoted the development of agriculture and improved crop yield and quality. For example, robot technology has enhanced the ability of skilled labor, greatly improving overall productivity in crop picking, weeding, sowing, and other areas \cite{verticalfarmingtemp3}.

\subsection{Drone-based and Robotic Pollination}

The pollination process is a relatively tedious step in agricultural cultivation \cite{nadeem2020review}. The new application of drones as pollination media deserves attention \cite{singh17yolov8}. Compared with other robot technologies, the entry threshold for drone technology is lower \cite{potts2018robotic}. The operation and movement of UAVs are controlled directly by pilots or autonomously moved through methods such as 3D environment modeling \cite{hiraguri2023autonomous}. Due to the different pollination methods required for crops, the application of pollination techniques for different drones also varies \cite{wu2024research}. It can be roughly divided into the following categories:
For example, the sowing type pollination adopted in the form of dry pollination for manual pollination of walnuts \cite{dronetemp1}, the liquid drop spray type adopted in the form of a water carrier for pollination of almonds and pears \cite{dronetemp2}, the airflow disturbance type shaken off by the built-in vibrator when pollinating tomatoes and strawberries \cite{tomato}, and the contact type delivered by a precision gun when pollinating sunflowers \cite{sunflower}.
\\\\
By bridging these gaps in robotic pollination, our system offers an integrated aerial platform for supporting contactless pollination. In this system, we combine a real-time custom-trained deep learning model for precise flower localization with a MPPI controller. This controller ensures stable flight dynamics and precise trajectory tracking even in the presence of the aerodynamic disturbances common in indoor farming.

\section{Proposed System}


The proposed system is an aerial manipulator platform designed for perception-driven flower targeting in vertical farming environments. It combines real-time visual perception, UAV flight control, and a lightweight manipulator to enable stable target approach and precise end-effector alignment near flowers. The current implementation focuses on positioning and alignment, while the modular end-effector interface supports future integration of contactless pollination mechanisms.

\subsection{System Overview}

\begin{figure}[!htbp]
    \centering \includegraphics[width=0.45\textwidth]{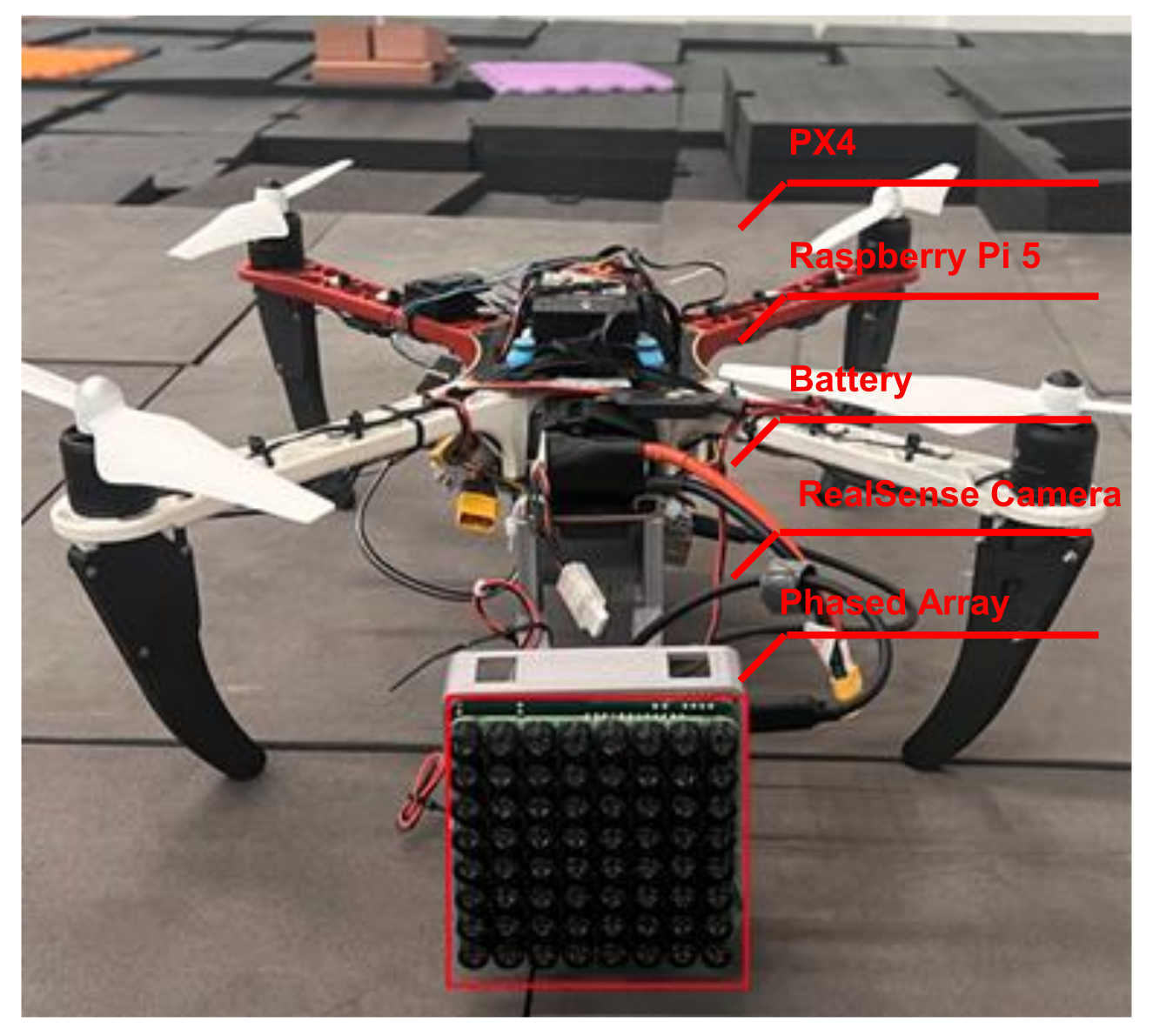}
    \caption{ 
    System overview of the proposed UAV aerial manipulator platform, integrating RGB-D-based flower detection, MPPI-based control on a PX4-enabled UAV, and a lightweight 2-DoF manipulator for precise end-effector positioning near target flowers in vertical farming environments.
    }
    \label{fig:SystemOverview}
\end{figure}

The platform is built on a lightweight quadrotor UAV integrated with onboard RGB-D perception, a lightweight 2-DoF manipulator, and a control stack based on the open-source PX4 autopilot. The system operates autonomously, with mission-level decisions driven by onboard perception and control algorithms that coordinate flower detection, target-relative positioning, and end-effector alignment, as shown in Figure \ref{fig:SystemOverview}.

\subsection{System Architecture}



\begin{figure*}[!htbp]
    \centering \includegraphics[width=\textwidth] {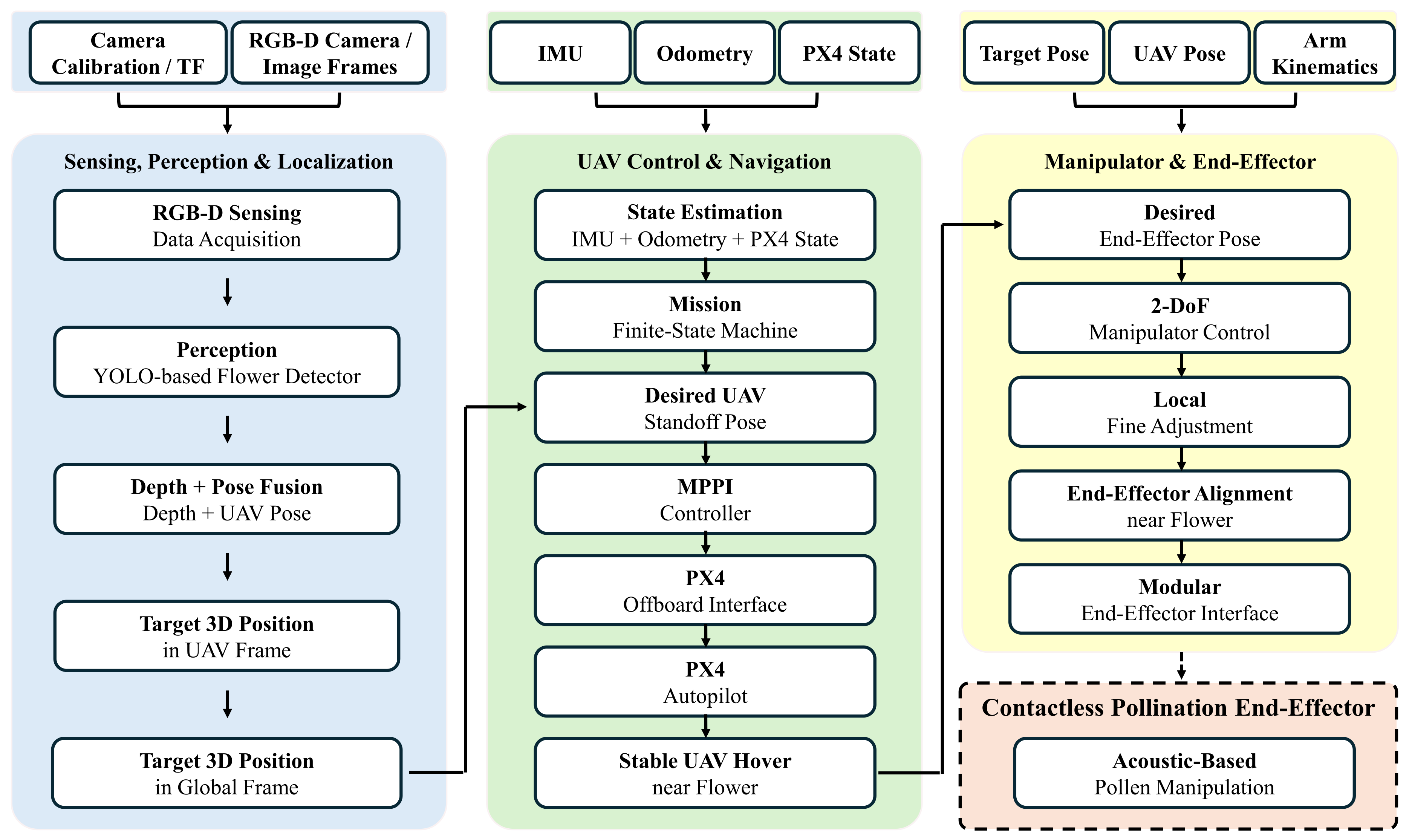}
    \caption{ 
    Modular system architecture of the proposed UAV aerial manipulator platform. 
    }
    \label{fig:SystemArchitecture}
\end{figure*}

The proposed system architecture adopts a modular design that enables real-time coordination among sensing, perception, localization, UAV control, and manipulator alignment components, as illustrated in Figure \ref{fig:SystemArchitecture}. 
The architecture is organized into three tightly coupled modules: \textit{Sensing, Perception \& Localization}, \textit{UAV Control \& Navigation}, and \textit{Manipulator \& End-Effector}. 


\subsubsection{Sensing, Perception \& Localization}



This module performs real-time flower sensing, detection, and localization using an onboard RGB-D camera system. The sensing layer first acquires RGB image frames and depth measurements, while camera calibration and frame transformation information are used to maintain geometric consistency between the camera, UAV, and global reference frames.
A lightweight deep learning detector, such as a YOLO-based flower detector trained on sunflower datasets, processes incoming image frames to identify flower regions in real time. The detected flower observations are then combined with depth measurements and UAV pose information through a depth-association and pose-fusion process. This enables the computation of the target 3D position first in the UAV frame and subsequently in the global frame. These target position estimates provide the spatial information required by the downstream control and manipulation modules for accurate aerial positioning and end-effector alignment near the flower \cite{zhao2025design}.

\subsubsection{UAV Control and Navigation}
This module is responsible for guiding the aerial platform to a stable standoff position near the detected flower. It receives IMU measurements, odometry, and PX4 state feedback, which are fused in the state estimation block to provide robust localization and motion feedback during indoor flight.
At the mission level, a finite-state machine governs the task progression through the main operational states, including search, detection, approach, alignment, and mission continuation. Once the flower target position is available from the perception and localization module, the system computes a desired UAV standoff pose that keeps the aerial platform at a safe operating distance from the flower while preserving the target within the sensing and manipulation workspace.
To realize this motion, the UAV uses an MPPI controller that generates closed-loop control actions under nonlinear dynamics and indoor disturbances. These commands are transmitted through the PX4 offboard interface to the PX4 autopilot, which executes the corresponding low-level flight control actions. As a result, the UAV achieves stable hover near the flower, thereby providing a reliable aerial base for the manipulator to perform precise local alignment \cite{enrico2025comparison}.

\subsubsection{Manipulator and End-Effector}


This module performs local positioning refinement once the UAV has reached a stable hover near the target flower. The desired end-effector pose is determined from the target pose, UAV pose, and manipulator kinematics, enabling accurate local adjustment relative to the flower.
A lightweight 2-DoF manipulator mounted beneath the UAV performs fine positioning corrections that complement the global mobility of the aerial platform. While the UAV provides coarse positioning and standoff stabilization, the manipulator improves targeting precision through local fine adjustment. This coordinated operation allows the system to achieve accurate end-effector alignment near the flower without requiring aggressive UAV maneuvers in constrained vertical farming environments.
The aligned end-effector is connected through a \textit{modular end-effector interface}, which is part of the current platform design and allows future integration of different pollination mechanisms. In the present implementation, the system focuses on perception-guided targeting, UAV standoff control, and precise end-effector positioning. Future extensions will incorporate contactless pollination end-effectors, including acoustic-based pollen manipulation modules.

\section{Design, Model, and Control}



This section describes the mechanical design, system modeling, and control algorithms of the proposed aerial manipulator platform. It focuses on achieving stable aerial positioning and precise end-effector alignment in constrained vertical farming environments.

\subsection{Mechanical Design}

\begin{figure}[!htbp]
    \centering \includegraphics[width=0.45\textwidth]{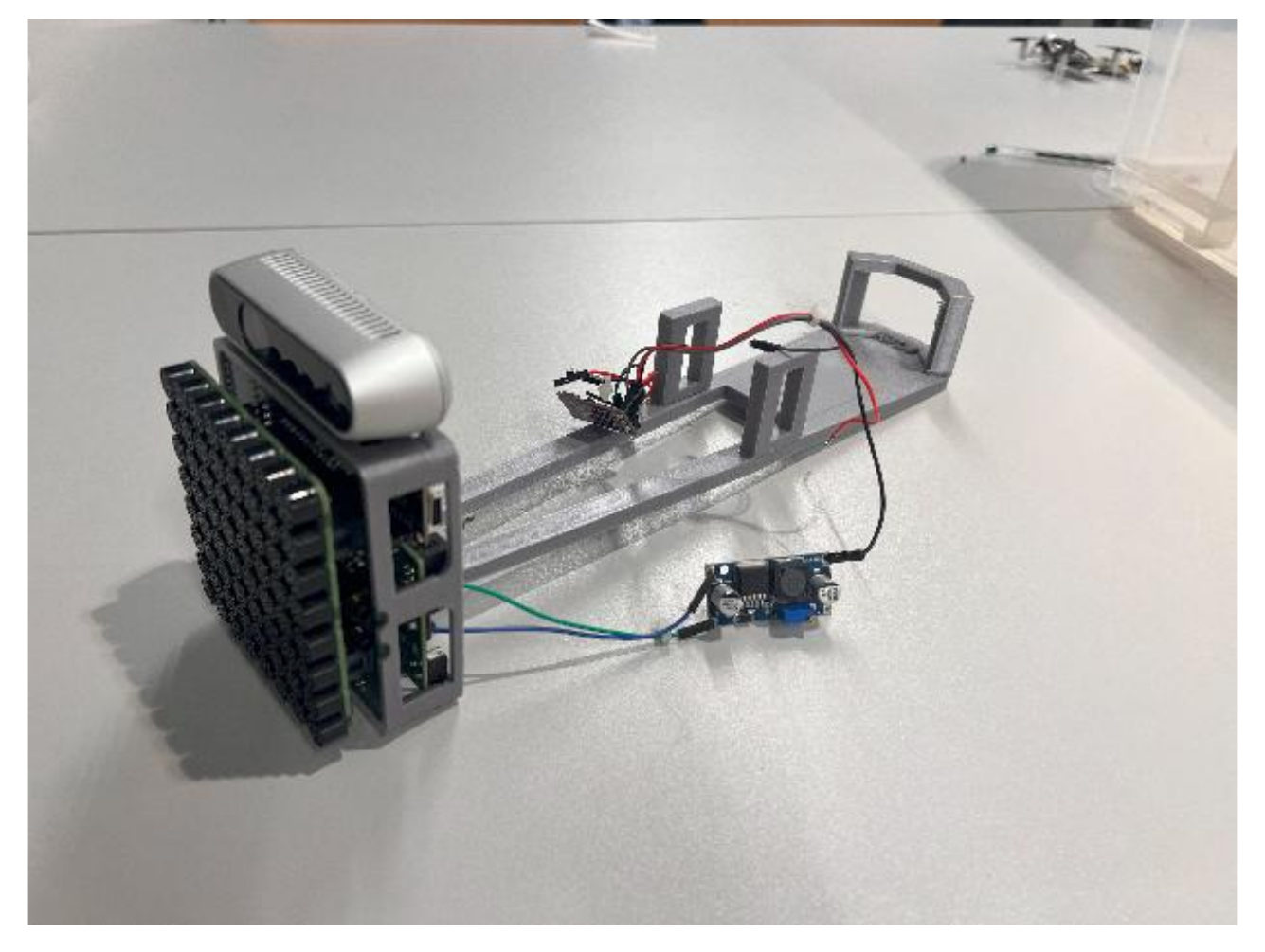}
    \caption{ 
    Mechanical design of the proposed UAV–manipulator platform, showing the custom L-shaped bracket for mounting the RGB-D camera and modular end-effector (phased array), enabling stable sensor alignment and precise positioning near target flowers.
    }\label{fig:MechanicalDesign}
\end{figure}

The core objective of this mechanical design is to support precise flower-targeted positioning in vertical farming, enabling the UAV to carry sensing and manipulation components from detection to target alignment. Instead of relying on direct contact, the platform uses a modular end-effector interface to support future study of contactless pollination mechanisms.
Sunflower pollination is cross-pollination, so traditional methods often use mechanical vibrators or brush-based contact. In contrast, this work explores a contactless approach using acoustic pollen transfer to the pistil stigma.
To support this functionality, we designed a custom L-shaped mounting structure to integrate the sensing and actuation components with the UAV platform. The bracket securely attaches the RGB-D camera and a modular end-effector interface beneath the drone, ensuring proper alignment with the target flower during operation. One end of the bracket is rigidly connected to the UAV frame. The other end provides a stable mounting point for the camera and the phased array as an end-effector.
%
%
The design emphasizes lightweight construction, structural stability, and minimal interference with UAV aerodynamics. It enables accurate positioning of sensing and actuation components relative to the target while maintaining flight stability. Figure \ref{fig:MechanicalDesign} shows the mechanical configuration and integration of the bracket with the UAV platform.

\subsection{UAV and Robotic Arm Modeling}



We designed a 2-DoF robotic arm integrated with the UAV platform to enable precise end-effector positioning near target flowers. The perception sensor (RGB-D camera) and a modular end-effector (phased-array) interface are mounted at the end of the manipulator, providing flexibility for integrating different actuation mechanisms in future extensions. 
%
The robotic arm is designed with a lightweight structure and is rigidly mounted beneath the UAV. While the manipulator itself provides only 2-DoF, the UAV provides full 6-DoF, enabling the combined system to achieve sufficient reachability and positioning capability within the workspace. This design balances mechanical simplicity and control effectiveness, allowing the manipulator to perform fine local adjustments while the UAV handles global positioning.
%
%
To model the system, the UAV–manipulator combination is treated as a coupled aerial manipulation system. The overall dynamics depend on the mass distribution, center of mass, and inertial properties of both the UAV and the attached arm. Key parameters considered in the model include the center of mass of the drone, the mass and geometry of the manipulator, joint angle limits, and the placement of the sensing and end-effector modules.


To ensure stable flight and accurate positioning, constraints are imposed on the manipulator motion, including limits on joint angles and actuation range. These constraints help maintain system stability by preventing excessive shifts in the UAV's center of mass and minimizing disturbances during operation. Additionally, the manipulator's influence on the UAV’s attitude and torque is accounted for in the control design, enabling coordinated motion between the UAV and the robotic arm.

\subsubsection{UAV Dynamics}

The UAV is modeled as a rigid-body quadrotor with 6-DoF. The translational and rotational dynamics are given by:
\begin{equation}
    m\ddot{p} = R f_T - m g
\end{equation}
\begin{equation}
    J\dot{\omega} = \tau - \omega \times (J\omega)
\end{equation}

where, $p \in \mathbb{R}^3$ is the UAV position in the inertial frame, and $\dot{p}$ and $\ddot{p}$ denote the velocity and acceleration, respectively. $R \in SO(3)$ is the rotation matrix from the body frame to the inertial frame, and $f_T \in \mathbb{R}^3$ is the total thrust vector expressed in the body frame. The parameter $m$ denotes the mass of the UAV, and $g = [0, 0, g]^T$ represents the gravitational acceleration vector with $g = 9.81~\mathrm{m/s^2}$.

In the rotational dynamics, $\omega \in \mathbb{R}^3$ is the angular velocity expressed in the body frame, $J \in \mathbb{R}^{3 \times 3}$ is the inertia matrix, and $\tau \in \mathbb{R}^3$ is the control torque vector generated by the propellers. The operator $\times$ denotes the vector cross product.

\subsubsection{UAV-Manipulator Coupling}

The addition of the manipulator introduces dynamic coupling between the UAV body and the robotic arm motion. The overall system can be modeled as:
\begin{equation}
M(q)\ddot{q} + C(q,\dot{q})\dot{q} + G(q) = u
\end{equation}

where $q \in \mathbb{R}^n$ represents the combined state vector of the UAV and manipulator, including UAV position, orientation, and manipulator joint angles. $M(q) \in \mathbb{R}^{n \times n}$ denotes the inertia matrix, $C(q,\dot{q})\dot{q} \in \mathbb{R}^n$ represents Coriolis and centrifugal effects, $G(q) \in \mathbb{R}^n$ is the gravity vector, and $u \in \mathbb{R}^n$ is the control input vector.

To reduce computational complexity, the manipulator is treated as a quasi-static subsystem, assuming relatively slow joint motion compared to UAV dynamics. This approximation enables a decoupled control strategy while maintaining sufficient accuracy for positioning and alignment tasks.




\subsubsection{End-Effector Kinematics}

The position of the end-effector is computed using forward kinematics:
\begin{equation}
x_e = p_{\mathrm{UAV}} + R_{\mathrm{UAV}} \, p_{\mathrm{arm}}
\end{equation}

where $x_e \in \mathbb{R}^3$ is the end-effector position in the inertial frame, $p_{\mathrm{UAV}} \in \mathbb{R}^3$ is the UAV position, and $R_{\mathrm{UAV}} \in SO(3)$ is the rotation matrix of the UAV. The vector $p_{\mathrm{arm}} \in \mathbb{R}^3$ denotes the position of the end-effector relative to the UAV body frame, which is determined by the manipulator joint angles through forward kinematics.

This formulation enables precise mapping between UAV pose, manipulator configuration, and the target flower position, facilitating accurate end-effector alignment during operation.




\subsection{UAV and Robotic Arm Control}

To execute feedback control for the UAV-manipulator system along the planned (or implicit) trajectory, we employ a MPPI control strategy \cite{belvedere2025feedback}. 

\subsubsection{MPPI-Based UAV Control}

The UAV control is implemented using a MPPI controller, which enables robust trajectory tracking under nonlinear dynamics and disturbances. 
At each control step, MPPI samples multiple control trajectories and minimizes the expected cost:
\begin{equation}
    \mathbf{u}^* = \arg \min \mathbb{E} \left[ \sum_{t=0}^{T} \ell(\mathbf{x}_t, \mathbf{u}_t) \right]
\end{equation}
where, the cost function $\ell(\cdot)$ penalizes: i) position error relative to the target, ii) velocity deviation, and iii) control effort.

The MPPI controller generates smooth control inputs that are sent to the PX4 autopilot through an offboard control interface.

\subsubsection{Robotic Arm Control}

The control objective is to maintain a stable UAV pose while aligning the end-effector with the desired flower position. 

The desired UAV position is computed as:
\begin{equation}
    \mathbf{p}_{des} = \mathbf{p}_{target} + \mathbf{d}_{offset}
\end{equation}
where, $\mathbf{d}_{offset}$ ensures a safe standoff distance from the flower ($\mathbf{p}_{target}$).

The manipulator performs fine adjustments to compensate for residual positioning errors, improving targeting precision without requiring aggressive UAV motion.
The control pipeline integrates perception, planning, and actuation: i) flower position is estimated by the perception module, ii) target coordinates are transformed to the UAV reference frame, iii) MPPI generates optimal UAV control inputs, iv) the manipulator adjusts end-effector alignment, and v) the system updates continuously based on feedback.
This hierarchical control strategy enables stable operation in cluttered indoor environments.

\section{Development}

%
%

\begin{figure*}[!htbp]
    \centering \includegraphics[width=\textwidth]{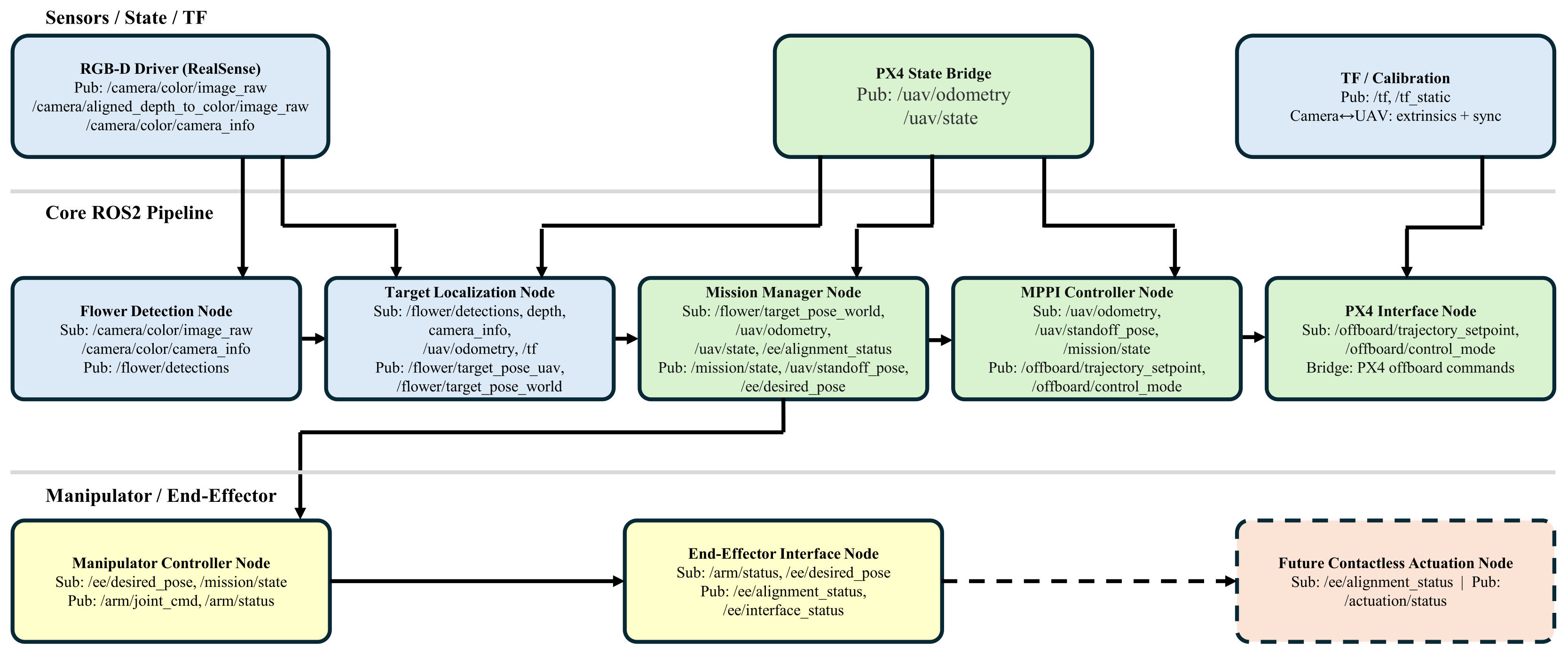}
    \caption{ 
    ROS 2-based software architecture of the proposed UAV aerial manipulator platform. RGB-D sensing, PX4 state feedback, and TF/calibration feed the perception, localization, control, and manipulator pipelines. The dashed node denotes an optional future contactless actuation extension.
    }
    \label{fig:SystemDevelopment}
\end{figure*}

The proposed system is implemented in ROS 2, enabling modular, real-time communication among sensing, perception, control, and manipulation components, as shown in Figure \ref{fig:SystemDevelopment}. 

The software stack consists of: i) sensor driver and state bridge nodes for RGB-D data and UAV feedback, ii) a perception and target-localization pipeline for flower detection and target estimation in the UAV frame, iii) a mission manager and MPPI-based control pipeline for trajectory generation and closed-loop UAV positioning through PX4, iv) a manipulator and end-effector interface pipeline for fine adjustment and alignment, and v) logging and visualization nodes for experiment monitoring and replay. 
The architecture is designed to support both MuJoCo simulation and the real-world PX4-based UAV platform using consistent ROS 2 interfaces, and future contactless actuation modules can be integrated through the modular end-effector interface.

All modules were implemented in Ubuntu 22.04 with ROS 2 Humble and deployed across both the MuJoCo simulation stack and the PX4-based experimental platform using unified launch files and parameter configurations. Hence, this design for MuJuCo simulation and the real-world PX4 environment preserves identical publishers, subscribers, nodes, and topics. As a result, algorithms proven in simulation are deployed to the real world by simply switching launch files and parameter YAMLs, while maintaining deterministic timing and safety behavior.
For more information on the modules developed, please refer to the GitHub repository\footnote{\url{https://github.com/Headmaster218/SkyGrip}}.

\section{Evaluation, Results, and Discussions}

This section evaluates the performance of the proposed aerial manipulator platform through both simulation and real-world experiments. The evaluation focuses on system stability, perception-guided positioning, and end-effector alignment accuracy in controlled environments. The goal is to validate the platform as a reliable testbed for future contactless pollination research.

\subsection{Simulation Experiments in MuJoCo}

\begin{figure}[!htbp]
    \centering \includegraphics[width=0.45\textwidth]{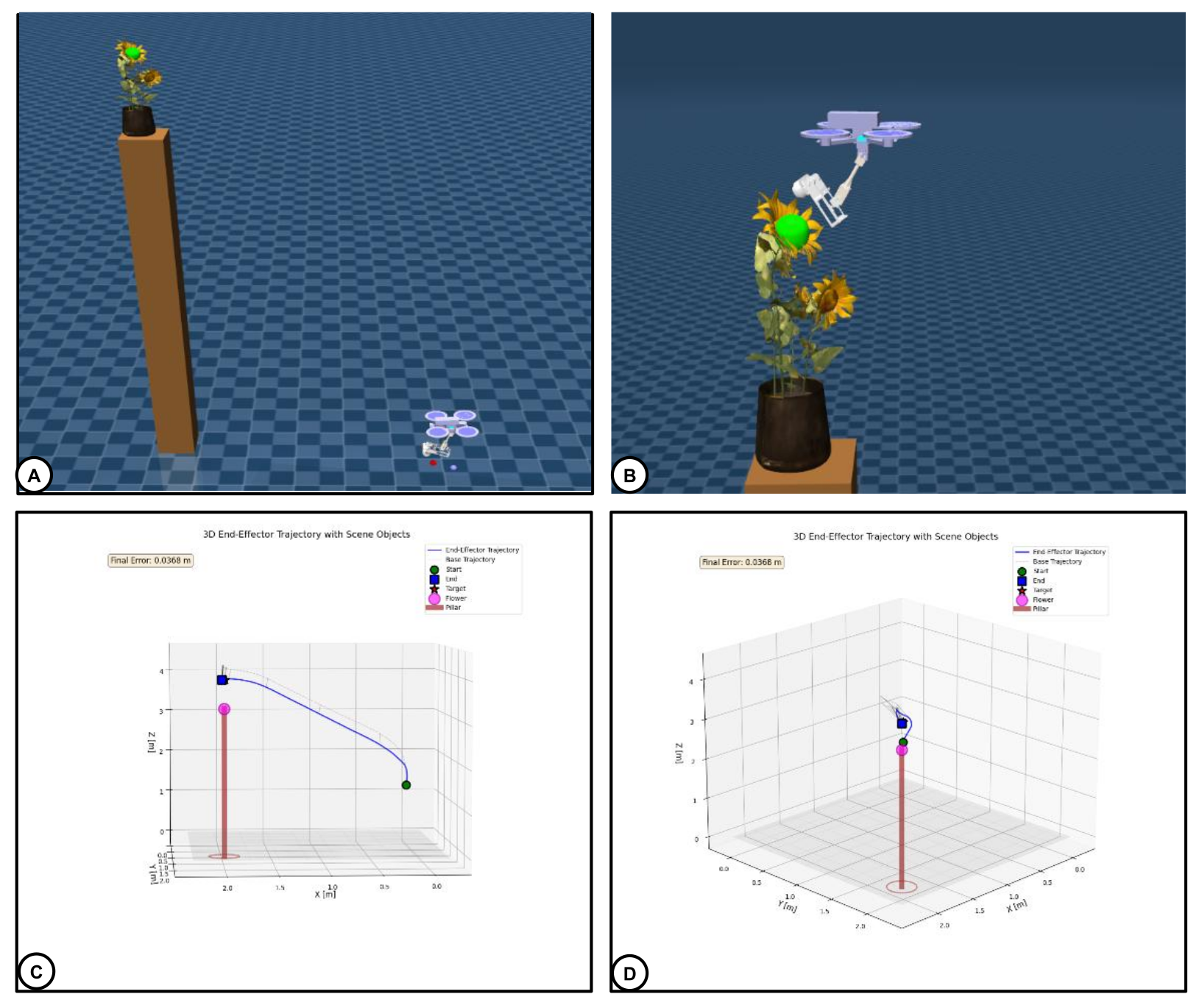}
    \caption{
    MuJoCo simulation for the UAV aerial manipulator system: A) simulation environment, B) UAV–manipulator interaction with a flower target.
    }
    \label{fig:SimulationResults}
\end{figure}

We first evaluate the system in a physics-based MuJoCo simulation environment to assess control performance and positioning accuracy under idealized conditions. The UAV aerial manipulator system is modeled at a 1:1 scale, including accurate geometry, mass distribution, and onboard sensing components. The simulation replicates the perception-to-control pipeline described in Section III and IV, enabling closed-loop evaluation of UAV motion and manipulator positioning.
The primary evaluation metric is the end-effector position error relative to the target flower. Orientation error is not modeled separately, as the current control objective prioritizes spatial alignment. A task is considered successfully aligned when the end-effector distance error is within 5 cm.
Simulation results demonstrate that the MPPI-based control framework enables stable trajectory tracking and consistent convergence toward the target position. The UAV maintains a stable hover while the manipulator performs fine adjustments, resulting in accurate end-effector placement across multiple trials (as shown in Figure \ref{fig:SimulationResults}).


\subsection{Real-World Experiments in UAV Lab}

\subsubsection{Experimental Setup}


Real-world experiments are conducted in a laboratory environment that mimics a vertical farming setup. The testbed is a 6 × 6 m indoor space, with sunflower targets arranged on vertical stands to replicate indoor vertical farm conditions.
The UAV operates autonomously using onboard perception and MPPI-based control on the PX4 platform. During operation, the UAV maintains a safe standoff distance of approximately 10–30 cm from the flower while aligning the end-effector with the target, as shown in Figure \ref{fig:RealWorldResults}.
To evaluate positioning performance, PhaseSpace motion capture markers and drivers are attached to both the end-effector and the target flowers, enabling accurate measurement of position and orientation deviations.

\subsubsection{Experimental Results}

The real-world experiments validate the feasibility of the proposed platform for perception-driven aerial positioning and manipulation tasks:
%
%
i) Perception and alignment: The perception module successfully detects and localizes flower targets in real time, enabling reliable alignment of the end-effector, 
%
%
ii) Positioning accuracy: The $\sim$40 mm difference along the X-axis matches the +40 mm forward standoff specified by the controller, showing the system was intentionally positioned in front of the flower as planned. Discrepancies in Y and Z reflect the lateral and vertical standoffs, as well as a slight axis misalignment between the coordinate frames,
%
%
3) Orientation error: Minor deviations in orientation are observed, which can be attributed to lighting conditions and sensing noise, and
iv) Trajectory Stability: The MPPI controller enables smooth and stable UAV motion, even in the presence of minor disturbances. The UAV maintains hover stability while the manipulator performs local adjustments, demonstrating effective coordination between aerial and manipulator subsystems.

\begin{figure}[!htbp]
    \centering \includegraphics[width=0.45\textwidth]{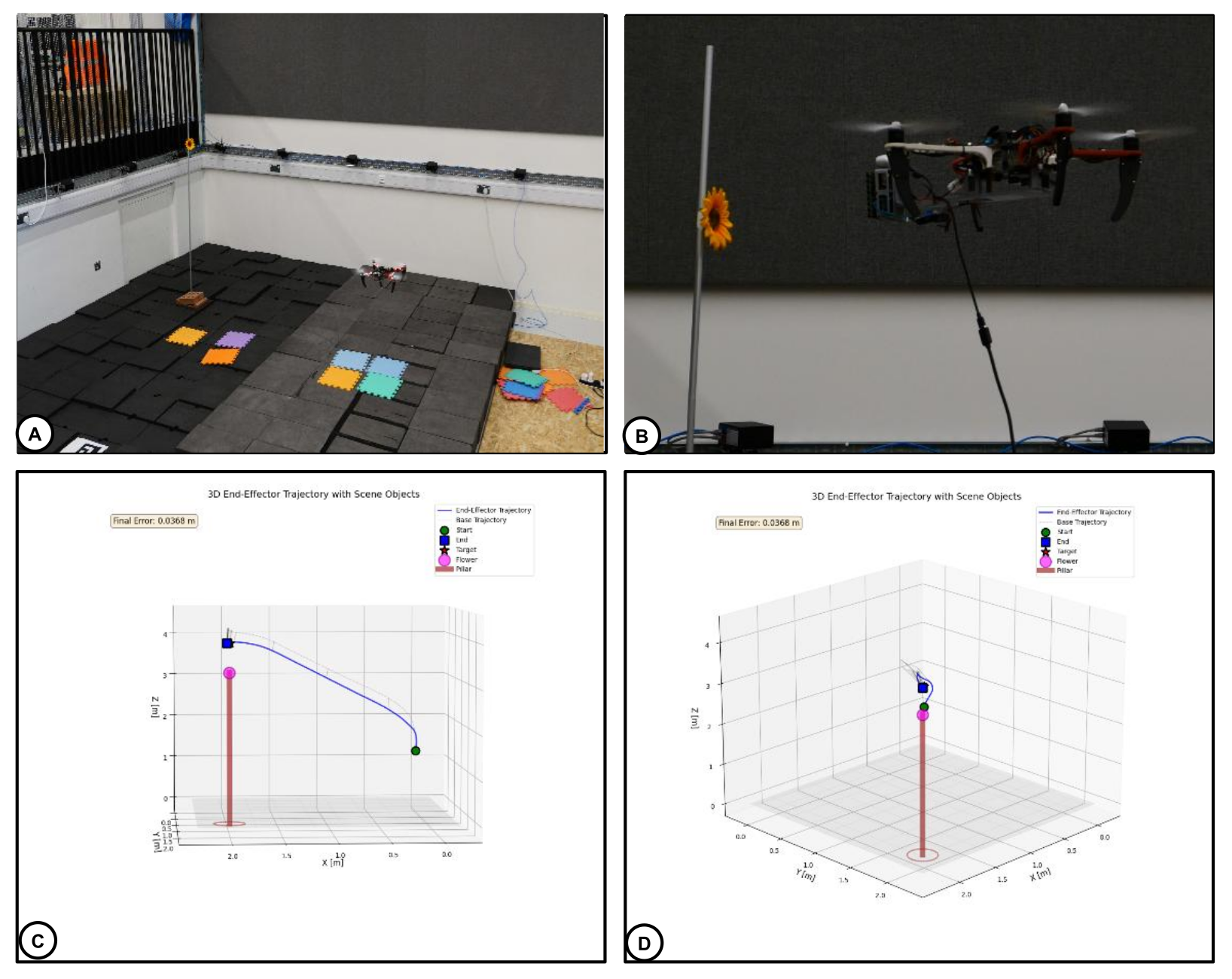}
    \caption{
    Real-world evaluation of the proposed UAV aerial manipulator platform: (A) indoor laboratory testbed replicating a vertical farming environment, (B) UAV–manipulator system performing flower-targeted positioning, (C) end-effector trajectory during approach, and (D) positioning accuracy relative to the target in 3D space.
    }
    \label{fig:RealWorldResults}
\end{figure}

Qualitative results from the experiments are shown in Figure \ref{fig:RealWorldResults}, illustrating stable UAV positioning and accurate end-effector alignment near the target flowers.

\subsection{Discussion}

The experimental results demonstrate that the proposed aerial manipulator platform provides a stable and accurate system for flower-targeted positioning in vertical farming environments.
The simulation and real-world evaluations confirm that: i) MPPI-based control framework enables robust aerial positioning, ii) perception system supports reliable flower localization, and iii) manipulator improves fine alignment accuracy without destabilizing the UAV.
Importantly, this work focuses on validating the robotic aerial manipulator and positioning framework, which is a critical prerequisite for future contactless pollination systems. While the current implementation demonstrates precise positioning near flower targets, integrating contactless actuation mechanisms, such as acoustic-based pollen manipulation using GS-PAT (Gerchberg-Saxton for Phased Arrays of Transducers) algorithm \cite{plasencia2020gs}, remains an important direction for future research.
Future work will investigate: i) improved perception robustness under varying lighting conditions, ii) enhanced calibration between sensing and actuation modules, and iii) integration of contactless end-effectors for controlled pollen delivery.

\section{Conclusion}

This paper presented an aerial manipulator platform for perception-driven flower targeting and precise end-effector positioning in vertical farming environments. The proposed system integrates onboard vision-based perception, MPPI-based UAV control, and a lightweight robotic manipulator to enable accurate and stable positioning under aerial mobility constraints.
The system was evaluated through MuJoCo simulations and real-world UAV lab experiments. The results show that the platform achieves stable flight, reliable flower localization, and centimeter-level end-effector positioning. These findings validate the effectiveness of the integrated perception–control–manipulation framework and highlight the feasibility of deploying aerial robotic systems in constrained agricultural environments.
Importantly, this work focuses on establishing a robust robotic carrier and positioning framework, which serves as a foundation for future contactless pollination technologies. While the current implementation emphasizes precise targeting and alignment, it provides the necessary infrastructure for integrating advanced end-effectors, such as acoustic-based pollen manipulation systems.
Future work will focus on improving perception robustness under varying environmental conditions \cite{cheng2025pollination}, enhancing calibration between sensing and actuation modules, and integrating contactless actuation mechanisms for controlled pollen delivery. These developments will further advance the realization of fully autonomous, contactless robotic pollination systems for sustainable indoor agriculture.





\bibliographystyle{unsrt} 
\bibliography{references}

\appendix

\end{document}